\documentclass{article}

\usepackage{arxiv}
\usepackage{amsmath}
\usepackage[utf8]{inputenc} 
\usepackage[T1]{fontenc}    
\usepackage{hyperref}       
\usepackage{url}            
\usepackage{booktabs}       
\usepackage{amsfonts}       
\usepackage{nicefrac}       
\usepackage{microtype}      
\usepackage{lipsum}
\usepackage{graphicx}
\graphicspath{ {./images/} }

\title{Enhancing Workplace Productivity and Well-Being using AI Agents}

\author{
 Dr Ravirajan K \\
  Associate Principal \\
  LTIMindtree \\
  USA \\
  \texttt{ravirajan.k@ltimindtree.com} \\
  \And
 Arvind Sundarajan \\
  Senior Director \\
  LTIMindtree \\
  Poland \\
  \texttt{arvind.sundararajan@ltimindtree.com} \\
  \texttt{} \\
}

\begin{document}
\maketitle
\begin{abstract}
This paper discussed about using Artificial Intelligence (AI) to enhance workplace productivity and employee well-being. By integrating  machine learning(ML) techniques with neurobiological data, the proposed approaches ensure alignment with human ethical standards through value alignment models and Hierarchical Reinforcement Learning (HRL) for autonomous task management. The system utilizes biometric feedback from the employees to generate personalized health prompts, fostering a supportive work environment that encourages physical activity. Additionally, we explore decentralized multi-agent systems for improved collaboration and decision-making frameworks that enhance transparency. Different approaches using ML techniques in hybrid with executing AI agents approaches are discussed. Together, these reviews aims to bring innovations and more productive and health-conscious workplace. AI agents accelerate these outcomes help the HR management and organization to launch the more rationale career progression stream for employees and organizational transformation.\end{abstract}


\section{Introduction}
The integration of Artificial Intelligence (AI) within contemporary organizational frameworks is increasingly recognized as a pivotal component in optimizing workplace productivity and fostering employee well-being. Employee engagement, attrition, carrer progression and employee satisfaction are the key pillars for any organizational transformation for a sustainable growth. Often, human resources management finds difficulty in customizing policies to sustain the workplace productivity. Several tangible and intangible factors are influencing the productivity and ofcourse it depends on the employees satisfaction. Conventional system based data would not support all the time to derive the appropriate policies, it is indeed to capture the intangible data such as cognitive metrics of individual employees will help to fine the policies for expected outcome and employee specific progression. In this rational, the papers attempts to engaged the digital assests such as AI- agent to analyse and provide the business support system for human resources management.

A comprehensive AI-driven system that encompasses a variety of tools designed to mitigate cognitive distractions, mental health, and enhance engagement through mechanisms such as adaptive gamification, intelligent task prioritization, and ambient health promotion. Central to this framework is a neuro-economic optimization model that synthesizes neuro-biological data, cognitive load metrics, and emotional state assessments to dynamically modulate workplace distractions. This model employs neuro-imaging technologies, including functional Magnetic Resonance Imaging (fMRI) and Electroencephalography (EEG), to formulate a constrained optimization problem.

The specific objectives is paramount in guiding research focused on the intersection of Artificial Intelligence (AI), workplace productivity, and employee well-being. This investigation aims to elucidate particular aspects such as cognitive load management, emotional state modulation, and the efficacy of adaptive gamification techniques. The primary hypothesis posits that the integration of AI-driven frameworks can significantly enhance productivity metrics while concurrently promoting mental health outcomes among employees. To systematically explore this hypothesis, the following research questions are proposed: 
a.How do neuroeconomic models inform the design of AI interventions that mitigate cognitive distractions? 
b.What is the impact of real-time performance analytics on task prioritization and employee engagement? 
Furthermore, the effectiveness of personalized health interventions can be quantitatively assessed through the econometric functions:

This paper is grounded in preliminary research that explores the integration of Artificial Intelligence (AI) within workplace environments, focusing on its potential to enhance productivity and employee well-being. The foundational analysis employs simulated data to model various scenarios and evaluate the effectiveness of the proposed AI-driven framework. By utilizing synthetic datasets, the research investigates key components such as cognitive load management, emotional state modulation, and adaptive gamification techniques. The simulations allow for the formulation of a constrained optimization problem.

 \section{Literature Review}
\label{sec:Literature Review}

The integration of Artificial Intelligence (AI) within workplace environments has garnered substantial attention in recent years, reflecting a growing recognition of its potential to enhance productivity and employee well-being. A comprehensive review of existing literature reveals a spectrum of applications, from intelligent task automation to personalized health interventions. Notable successes include the implementation of adaptive gamification techniques that leverage real-time performance analytics to engage employees effectively, as evidenced by studies demonstrating improved task completion rates and heightened employee satisfaction. For instance, research has shown that gamified systems can lead to a significant increase in engagement metrics, thereby fostering a more productive work environment.

However, limitations persist within the literature. Many studies have focused narrowly on isolated AI applications without adequately addressing the complex interplay of neuroeconomic models, cognitive load management, and affective intelligence. This lack of holistic approaches may obscure the multifaceted nature of workplace dynamics influenced by AI technologies. Furthermore, empirical evidence supporting the long-term effectiveness of AI interventions remains sparse, with many studies relying on short-term evaluations that do not account for sustained behavioral changes over time.

Recent advancements in AI methodologies further enrich this discourse. Innovations such as Multi-Objective Reinforcement Learning (MORL) and Hierarchical Reinforcement Learning (HRL) have emerged as powerful tools for optimizing decision-making processes in dynamic environments. These frameworks allow for the simultaneous consideration of multiple conflicting objectives, enhancing the adaptability of AI systems in real-world applications. Additionally, developments in Explainable AI (XAI) have begun to address transparency concerns by providing interpretable insights into AI decision-making processes, thereby fostering trust among users.

In summary, while the literature presents compelling evidence of the benefits associated with AI integration in workplace settings, it also highlights critical gaps that warrant further exploration. Future research should aim to adopt a more comprehensive perspective that encompasses the intricate interactions between various AI components and their impact on employee productivity and well-being.

 \section{Data and Methodology}
 \subsection{Data}
\label{sec:Data}
The reliance on simulated data enables the exploration of various configurations and parameters within the framework without the constraints associated with real-world data collection. This approach facilitates a rigorous examination of theoretical models that underpin AI applications in organizational settings. Future research will aim to validate these findings through empirical studies involving actual workplace data, thereby enhancing the robustness of the conclusions drawn from this initial investigation. The insights gained from this foundational work are intended to inform subsequent phases of research that will incorporate real-world applications and further refine the proposed AI-driven solutions.

 \subsection{Methodology}
\subsubsection{Personalized Health}

The effectiveness of personalized health improvement can be derived using following function as below
\begin{equation}
\min_{\mathbf{x}} \quad f(\mathbf{x}) \quad \text{subject to} \quad g_i(\mathbf{x}) \leq 0, \quad h_j(\mathbf{x}) = 0,
\end{equation}

where \( f(\mathbf{x}) \) represents distraction variables to be minimized while maximizing cognitive engagement indices. An adaptive gamification subsystem is integrated into the framework, leveraging real-time performance analytics and sentiment analysis facilitated by AI algorithms. This customization of game mechanics for individual employee profiles can be expressed as:

\begin{equation}
G_i = \mathcal{F}(P_i, S_i),
\end{equation}

where \( G_i \) denotes the game mechanics for employee \( i \), \( P_i \) is the performance data derived from AI-driven analytics, and \( S_i \) is the sentiment analysis output generated through Natural Language Processing (NLP). The framework further incorporates intelligent task prioritization algorithms utilizing machine learning techniques to dynamically allocate and schedule tasks based on real-time assessments of employee workload and availability. The scheduling optimization can be formulated as:

\begin{equation}
\max_{T_{ijk}} \sum_{i,j,k} w_{ijk} T_{ijk},
\end{equation}

where \( w_{ijk} \) represents the weight assigned to each task based on its priority as determined by predictive analytics. Moreover, ambient health promotion techniques utilize biometric feedback and engagement metrics processed through AI models to identify and disseminate content conducive to employee well-being, modeled as:

\begin{equation}
C_i = H(P_i, E_i),
\end{equation}

where \( C_i \) is the content delivered to employee \( i \), and \( H(P_i, E_i) \) represents a function combining health prompts based on performance data and engagement metrics. The proposed system also incorporates an advanced AI agent framework where Value Alignment Models utilize inverse reinforcement learning to ensure alignment with human ethical standards. Multi-Objective Reinforcement Learning (MORL) facilitates the optimization of conflicting objectives while Hierarchical Reinforcement Learning (HRL) decomposes complex tasks into manageable hierarchies, enhancing autonomous operational capabilities. 
\begin{figure}[htbp] 
    \centering
    \includegraphics[width=0.45\textwidth]{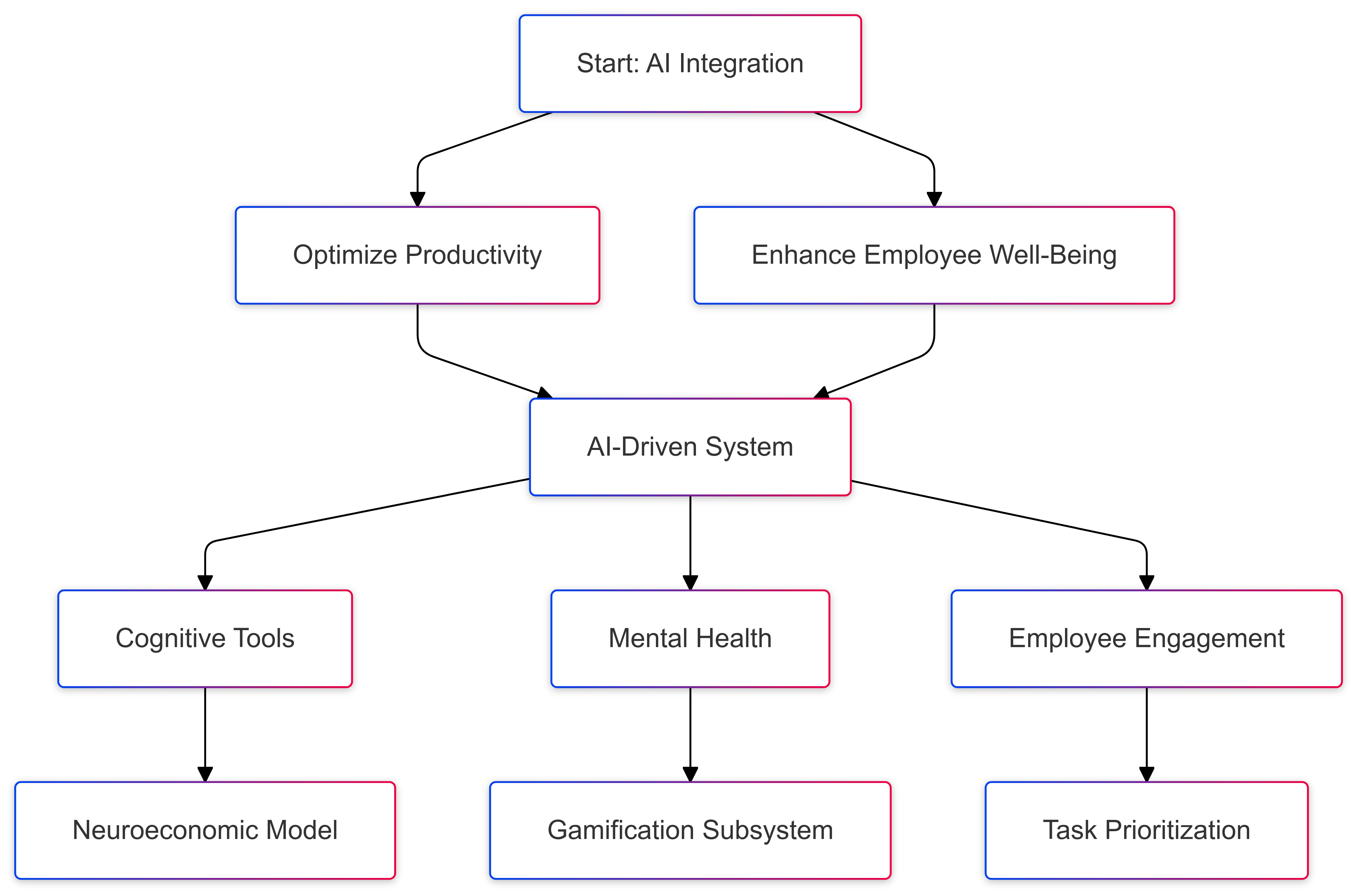}
    \caption{Overview of Employee Wellness AI Framework}
    \label{fig:Employee Wellness}
\end{figure}

In terms of decision-making, Chain-of-Thought prompting encourages agents to articulate their reasoning processes in a structured manner, complemented by Behavior Trees that define decision-making hierarchies. Lifelong Learning frameworks are integrated to enable agents to continuously adapt while retaining knowledge from prior tasks.

Finally, the architecture of the proposed system incorporates Explainable AI (XAI) frameworks, which enhance the transparency of decision-making processes, thereby facilitating trust and usability in AI applications within workplace settings.

\begin{equation}
H(P,E) = w_1 P + w_2 E,
\end{equation}

where \( H(P,E) \) represents the effectiveness of health interventions, \( P \) denotes physiological parameters (e.g., heart rate variability), and \( E \) indicates environmental factors (e.g., ambient noise levels). The weights \( w_1 \) and \( w_2 \) are empirically derived to reflect the significance of each parameter based on extensive data analysis. By establishing these clear objectives and hypotheses, this research endeavors to contribute to a nuanced understanding of how AI can be leveraged to optimize workplace environments, ultimately leading to enhanced organizational performance and improved employee well-being.

\begin{equation}
\min_{\mathbf{x}} \quad f(\mathbf{x}) \quad \text{subject to} \quad g_i(\mathbf{x}) \leq 0, \quad h_j(\mathbf{x}) = 0,
\end{equation}

where \( f(\mathbf{x}) \) encapsulates distraction variables targeted for minimization while maximizing cognitive engagement indices. Through this methodical approach, the study assesses the potential impacts of AI interventions on workplace dynamics, utilizing algorithms such as Multi-Objective Reinforcement Learning (MORL) and Hierarchical Reinforcement Learning (HRL) to optimize decision-making processes.

\subsubsection{Neuroeconomic Models in AI}

Neuroeconomic models leverage  statistical methods to analyze decision-making processes influenced by neural mechanisms, integrating neurobiological data, cognitive load metrics, and emotional state assessments to optimize workplace productivity. The optimization framework can be defined as:

\begin{equation}
V(x) = E[R(x)] - C(x),
\end{equation}

where \( V(x) \) represents the value of action \( x \), \( E[R(x)] \) is the expected reward from taking action \( x \), and \( C(x) \) denotes the associated cost. Inputs to this model include neuroimaging data from functional Magnetic Resonance Imaging (fMRI) and Electroencephalography (EEG), cognitive load assessments derived from performance metrics, and emotional evaluations obtained through sentiment analysis. The processing involves constructing a complex dataset that synthesizes these inputs to formulate a constrained optimization problem expressed as:

\begin{equation}
\min_{\mathbf{x}} \quad f(\mathbf{x}) \quad \text{subject to} \quad g_i(\mathbf{x}) \leq 0, \quad h_j(\mathbf{x}) = 0,
\end{equation}

where \( f(\mathbf{x}) \) encapsulates distraction variables to be minimized while maximizing cognitive engagement indices. Outputs from this framework include optimized decision strategies that enhance employee performance and real-time feedback mechanisms that adjust task allocations based on ongoing assessments of cognitive states.

\begin{figure}[htbp] 
    \centering
    \includegraphics[width=0.45\textwidth]{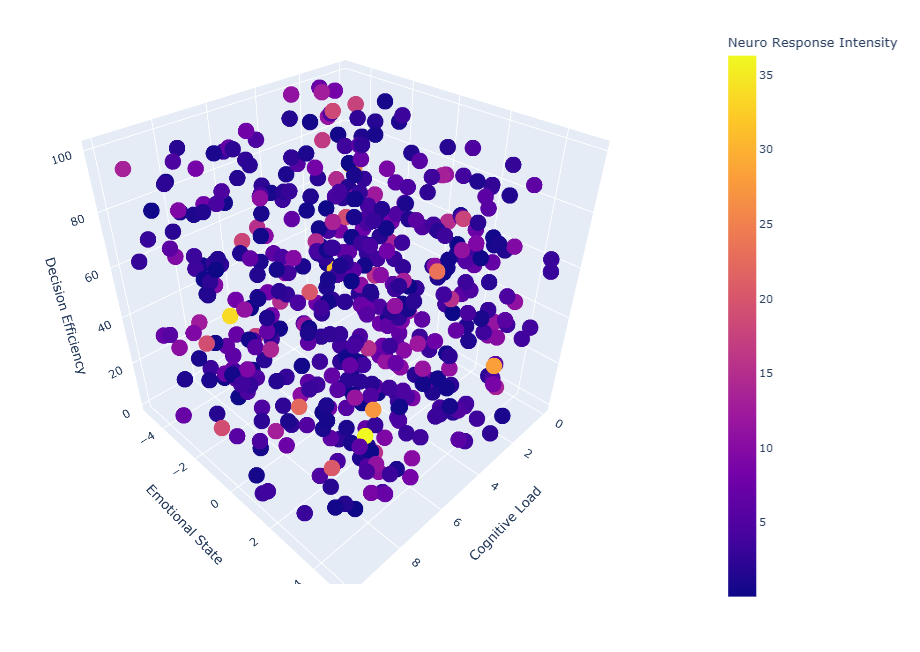}
    \caption{This 3D scatter plot visualizes cognitive load, emotional state, and decision efficiency metrics as part of a neuroeconomic model. Data points represent tasks or distractions, with color intensity indicating neural response levels}
    \label{fig:Employee Wellness}
\end{figure}

The key variables are: Cognitive Load, Emotional State, Decision Efficiency, Neuro Response, and Task Category where \( V(x) \) denotes the value of action \( x \), \( E[R(x)] \) is the expected reward associated with action \( x \), and \( C(x) \) represents the incurred costs. Inputs to this model include neuroimaging data from functional Magnetic Resonance Imaging (fMRI) and Electroencephalography (EEG), cognitive load assessments derived from performance tasks, and emotional evaluations obtained through sentiment analysis. The processing pipeline involves constructing a complex dataset that integrates these inputs to formulate a constrained optimization problem where \( f(\mathbf{x}) \) encapsulates distraction variables targeted for minimization while maximizing cognitive engagement indices. Outputs from this framework include optimized decision strategies that enhance task performance and real-time feedback mechanisms that adjust based on ongoing assessments of cognitive states. By employing reinforcement learning algorithms, the system iteratively refines its recommendations based on user interactions, thereby enhancing personalization and efficacy in mitigating cognitive distractions.

By employing reinforcement learning algorithms, specifically utilizing techniques such as Q-learning and Policy Gradient methods, the system continuously refines its recommendations based on user interactions, thereby enhancing personalization. The learning process can be formalized as:

\begin{equation}
Q(s,a) \leftarrow Q(s,a) + \alpha \left[ r + \gamma \max_{a'} Q(s',a') - Q(s,a) \right],
\end{equation}

where \( Q(s,a) \) is the action-value function for state \( s \) and action \( a \), \( r \) is the immediate reward, \( \alpha \) is the learning rate, and \( \gamma \) is the discount factor for future rewards. This iterative refinement allows for the adaptation of decision strategies that optimize cognitive engagement while minimizing distractions. Inputs to this model include real-time neuroimaging data from functional Magnetic Resonance Imaging (fMRI) and Electroencephalography (EEG), cognitive load assessments derived from performance tasks, and emotional evaluations obtained through sentiment analysis. 
\begin{figure}[htbp] 
    \centering
    \includegraphics[width=0.45\textwidth]{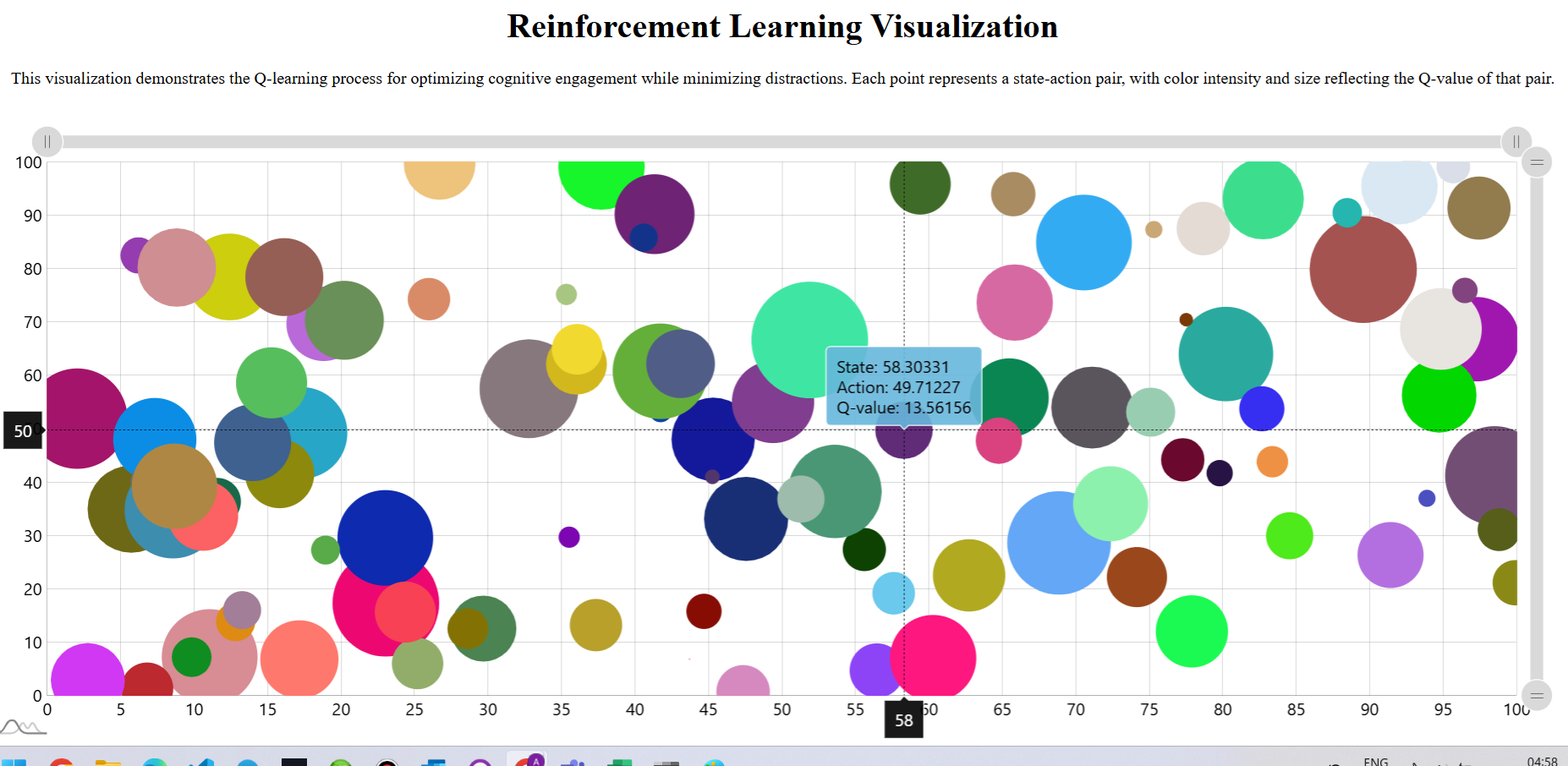}
    \caption{This visualization demonstrates the Q-learning process for optimizing cognitive engagement while minimizing distractions. Each point represents a state-action pair, with color intensity and size reflecting the Q-value of that pair}
    \label{fig:image2}
\end{figure}

Outputs encompass optimized decision strategies that enhance task performance and real-time feedback mechanisms that adjust based on ongoing assessments of cognitive states. By leveraging these methodologies, the framework aims to mitigate cognitive distractions effectively while promoting sustained employee engagement.

\subsubsection{Adaptive Gamification and Intelligent Task Prioritization}

The integration of an adaptive gamification subsystem within the AI framework leverages real-time performance analytics and sentiment analysis to dynamically customize game mechanics tailored to individual employee profiles. This customization can be modeled using reinforcement learning principles, specifically through the action-value function defined as:

\begin{equation}
Q(s, a) = R(s, a) + \gamma \max_{a'} Q(s', a'),
\end{equation}

where \( Q(s, a) \) represents the action-value function for state \( s \) and action \( a \), \( R(s, a) \) is the immediate reward received after executing action \( a \), and \( \gamma \) is the discount factor for future rewards. The adaptive nature of this system facilitates enhanced task prioritization by aligning game elements with individual performance metrics, thereby optimizing engagement and productivity. The framework employs machine learning algorithms to assess real-time data inputs, including cognitive load metrics and emotional state assessments, which are processed to inform task allocation strategies. The task scheduling optimization can be expressed as:

\begin{equation}
\max_{T_{ijk}} \sum_{i,j,k} w_{ijk} T_{ijk},
\end{equation}

where \( w_{ijk} \) denotes the weight assigned to each task based on its priority derived from predictive analytics. Through this iterative process, the system continuously refines its recommendations based on user interactions, enhancing personalization and efficacy in mitigating workplace distractions.
\begin{figure}[htbp] 
    \centering
    \includegraphics[width=0.45\textwidth]{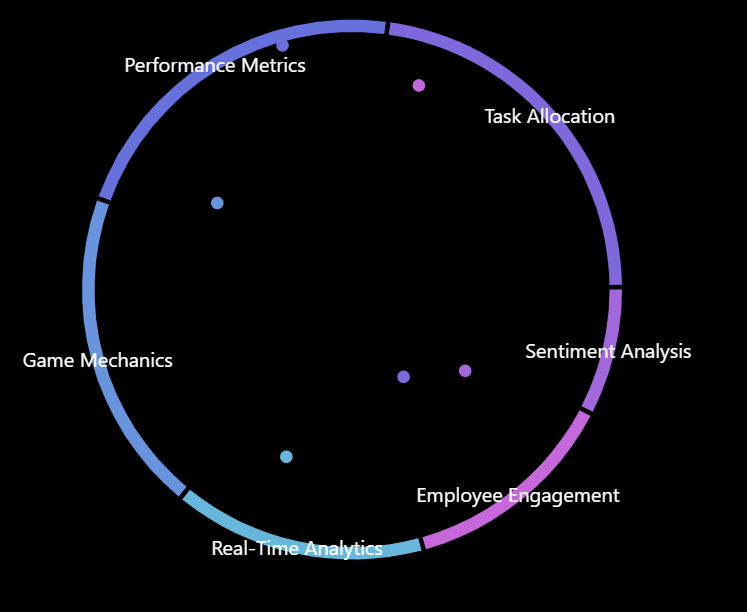}
    \caption{This chord diagram illustrates the interconnected elements of adaptive gamification and task prioritization,leveraging reinforcement learning principles to optimize engagement and productivity}
    \label{fig:image2}
\end{figure}

The primary function of the chord diagram is to facilitate the exploration of these relationships by dynamically rendering connections based on user-defined parameters. The effectiveness of this visualization can be modeled using reinforcement learning principles where \( Q(s, a) \) denotes the action-value for state \( s \) and action \( a \), \( R(s, a) \) represents the immediate reward from executing action \( a \), and \( \gamma \) is the discount factor for future rewards. 

This dynamic adjustment enhances task prioritization by aligning game elements with individual performance metrics. User interactions with the diagram trigger real-time updates to the visual representation, allowing for an iterative exploration of data that reflects ongoing changes in cognitive load and emotional states. The integration of these visual analytics tools underscores the role of AI in enhancing decision-making processes within organizational frameworks.

The incorporation of advanced algorithms such as Multi-Objective Reinforcement Learning (MORL) allows for simultaneous optimization of conflicting objectives, while Hierarchical Reinforcement Learning (HRL) decomposes complex tasks into manageable subtasks, facilitating autonomous decision-making capabilities in dynamic environments.

\subsubsection{Ambient Health Promotion Techniques}
Ambient health promotion techniques leverage biometric feedback to create personalized health interventions aimed at enhancing employee well-being. The effectiveness of these interventions can be quantified using a function that integrates physiological data, defined as:

\begin{equation}
H(P,E) = w_1 P + w_2 E,
\end{equation}

where \( H(P,E) \) represents the health intervention effectiveness, \( P \) denotes physiological parameters (e.g., heart rate variability, galvanic skin response), \( E \) indicates environmental factors (e.g., noise levels, lighting conditions), and \( w_1, w_2 \) are weights assigned based on empirical evidence derived from machine learning models. This model underscores the importance of contextual factors in promoting employee well-being and can be enhanced through reinforcement learning frameworks that adaptively tune the weights based on real-time feedback.

\begin{figure}[htbp] 
    \centering
    \includegraphics[width=0.45\textwidth]{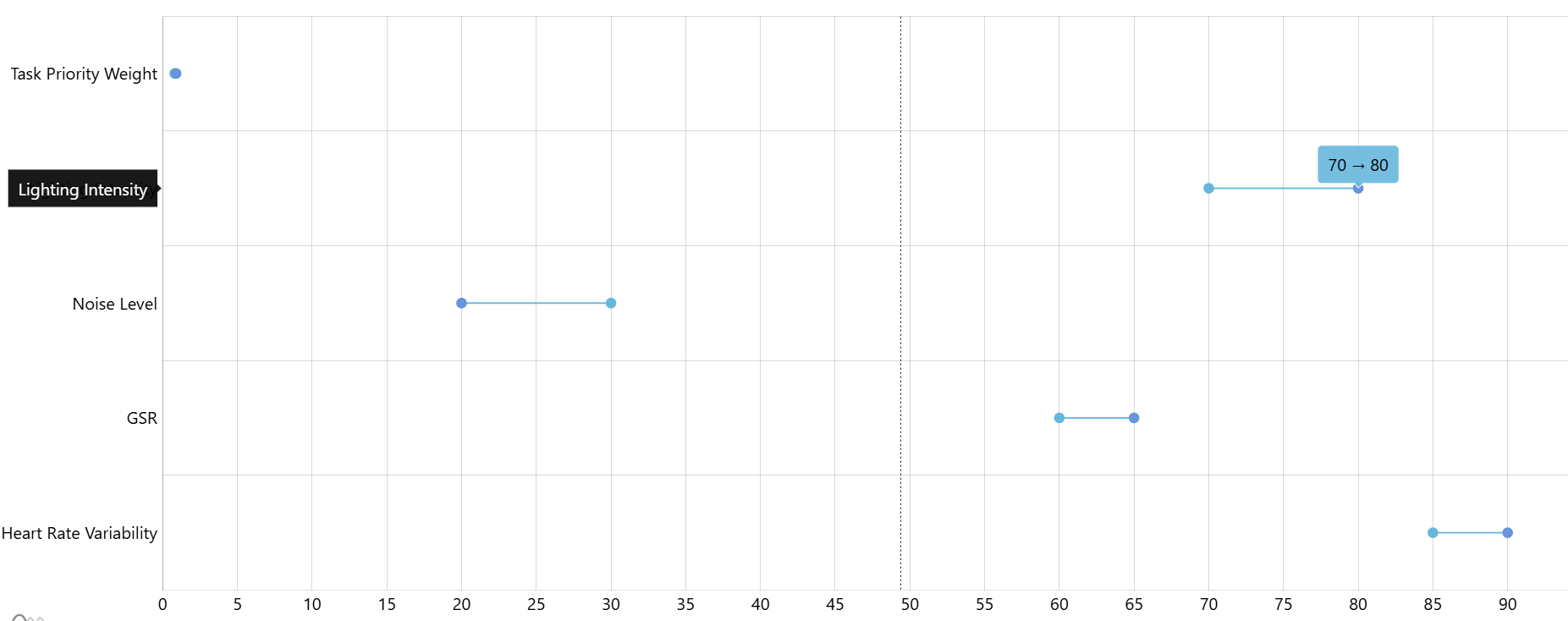}
    \caption{This visualization demonstrates the Q-learning process for optimizing cognitive engagement and promoting employee well-being. Each point represents a state-action pair, with color intensity and size reflecting the Q-value of that pair, showcasing how health interventions dynamically adapt to real-time biometric and environmental data.}
    \label{fig:image4}
\end{figure}

The integration of advanced AI methodologies facilitates the continuous monitoring and analysis of biometric data through wearable devices, enabling the system to dynamically adjust health interventions. The adaptive nature of this system can be modeled using reinforcement learning principles, wherein the action-value function is defined as:

\begin{equation}
Q(s, a) = R(s, a) + \gamma \max_{a'} Q(s', a'),
\end{equation}

where \( Q(s, a) \) denotes the action-value for state \( s \) and action \( a \), \( R(s, a) \) represents the immediate reward from executing action \( a \), and \( \gamma \) is the discount factor for future rewards. This iterative learning process allows for improved decision-making regarding health interventions by aligning them with individual performance metrics and emotional states.

Furthermore, ambient health promotion techniques utilize machine learning algorithms such as Support Vector Machines (SVMs) and Random Forests to classify physiological data patterns and predict potential health risks. The framework incorporates intelligent task prioritization algorithms that utilize real-time assessments of employee workload and availability to allocate tasks effectively. This can be expressed mathematically as:

\begin{equation}
\max_{T_{ijk}} \sum_{i,j,k} w_{ijk} T_{ijk},
\end{equation}

where \( w_{ijk} \) represents the weight assigned to each task based on its priority determined by predictive analytics.

By employing these advanced methodologies, the proposed system not only enhances workplace productivity through optimized task management but also supports employee mental health by delivering timely and contextually relevant health interventions. The synthesis of these approaches highlights significant advancements in understanding how AI can effectively promote well-being within organizational settings.

\subsubsection{AI Agent Capabilities and Frameworks}
The integration of Artificial Intelligence (AI) agents within organizational systems is pivotal for enhancing operational efficiency and decision-making processes. AI agents exhibit diverse capabilities, including autonomous operation, multi-agent collaboration, and adaptive learning strategies. 
\begin{table}
 \caption{Simulation on Synthetic data for Corporate Analysis of Productivity Scores and Key Findings}
  \centering
  \begin{tabular}{|l|l|l|l|l|}
    \hline
    \textbf{Group} & \textbf{Age Range} & \textbf{Gender} & \textbf{Productivity Score (1-5)} & \textbf{Key Findings} \\
    \hline
    Group A & 18-25 & Male & 4.2 & Higher engagement in collaborative tasks. \\
    Group B & 18-25 & Female & 4.5 & Excels in tasks requiring creativity and multitasking. \\
    Group C & 26-35 & Male & 3.8 & Demonstrates consistent performance in technical roles. \\
    Group D & 26-35 & Female & 4.3 & Strong in strategic planning and decision-making tasks. \\
    Group E & 36-45 & Male & 3.9 & Prefers independent work over group collaborations. \\
    Group F & 36-45 & Female & 4.0 & Shows balanced performance across multiple domains. \\
    Group G & 46-55 & Male & 3.7 & Demonstrates steady but slower adaptability to new tools. \\
    Group H & 46-55 & Female & 4.1 & Excels in mentoring and team management. \\
    Group I & 56+ & Male & 3.5 & Focused expertise but reduced multitasking capabilities. \\
    Group J & 56+ & Female & 3.9 & Contributes effectively in advisory and quality-check roles. \\
    \hline
  \end{tabular}
  \label{tab:Demographic_Productivity_Analysis}
\end{table}

Autonomous agents are designed to function independently, making decisions based on real-time data inputs without human intervention. This autonomy is facilitated through advanced algorithms such as Multi-Objective Reinforcement Learning (MORL), which allows agents to optimize multiple conflicting objectives simultaneously, ensuring alignment with human values.

\begin{figure}[htbp] 
    \centering
    \includegraphics[width=0.45\textwidth]{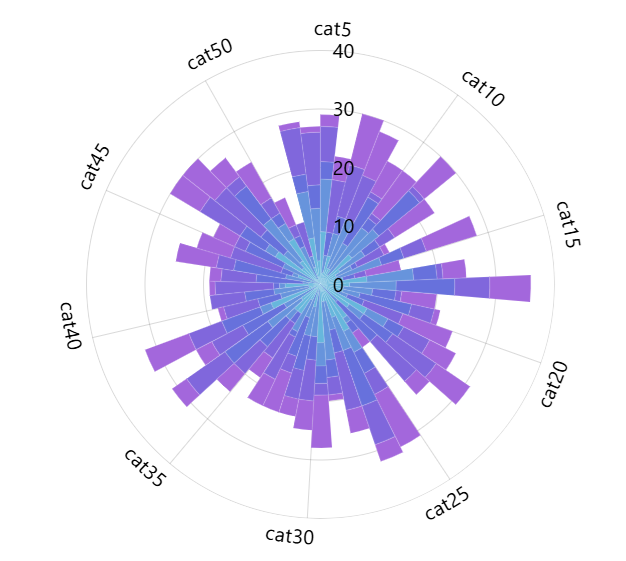}
    \caption{This visualization demonstrates the integration of AI agent capabilities within organizational systems, focusing on multi-agent collaboration, adaptive learning, and real-time health interventions. Each axis represents a different AI feature such as Autonomous Operation, Generative Adversarial Networks, Biometric Feedback, and more. The chart showcases how AI agents can optimize task allocation, decision-making, and employee well-being by dynamically adjusting based on real-time biometric and environmental data. The varying color intensity and size of each data point reflect the performance and alignment with human values in real-world applications.}
    \label{fig:image5}
\end{figure}

In the realm of multi-agent systems, agents engage in structured debates using argumentation frameworks that enhance their reasoning capabilities. This interaction is critical for developing robust decision-making models where agents can simulate human-like behavior through Generative Adversarial Networks (GANs). The action selection process is governed by sophisticated algorithms such as Deep Q-Networks (DQN) that utilize neural networks for efficient learning from experience. The action-value function can be expressed as:

\begin{equation}
Q(s, a) = R(s, a) + \gamma \max_{a'} Q(s', a'),
\end{equation}

where \( Q(s, a) \) represents the expected utility of taking action \( a \) in state \( s \), \( R(s, a) \) denotes the immediate reward received, and \( \gamma \) is the discount factor for future rewards.

Active perception mechanisms enable agents to strategically gather information from their environment, enhancing their understanding and adaptability.

These mechanisms are supported by attention models that prioritize relevant features, thereby improving decision quality. Additionally, the framework incorporates hierarchical reinforcement learning (HRL), which decomposes complex tasks into manageable subtasks, allowing efficient planning and execution.

The implementation of ambient health promotion techniques further exemplifies the application of AI agents to improve employee well-being. Using biometric feedback and engagement metrics processed through AI models, organizations can deliver personalized health interventions that adapt to individual needs. The effectiveness of these interventions can be quantified using the following function.

\begin{equation}
H(P,E) = w_1 P + w_2 E,
\end{equation}

where \( H(P,E) \) signifies the effectiveness of health intervention, \( P \) denotes physiological parameters such as heart rate variability, \( E \) indicates environmental factors such as noise levels and \( w_1, w_2 \) are weights determined by empirical analysis.

Through these advanced methodologies, AI agents not only optimize task allocation and enhance collaboration, but also contribute significantly to promoting mental health in workplace environments. The systematic integration of these capabilities underscores the transformative potential of AI in contemporary organizational frameworks.

\section{Use Cases and Business Potential}
The implementation of an AI-driven framework that integrates neuroeconomic principles, adaptive gamification, and intelligent task prioritization presents significant use cases across various sectors, including corporate environments, healthcare, and education. In corporate settings, the framework can enhance employee productivity by dynamically modulating cognitive distractions through real-time analytics derived from biometric feedback. This can be modeled as:

\begin{equation}
D = f(C, E),
\end{equation}

where \( D \) represents distraction levels, \( C \) denotes cognitive load metrics, and \( E \) indicates environmental factors. By leveraging reinforcement learning algorithms to optimize task allocation based on individual workload assessments, organizations can achieve substantial improvements in operational efficiency.

In healthcare, the ambient health promotion techniques embedded within the framework facilitate personalized interventions that adapt to individual physiological states. The effectiveness of these interventions can be quantified using:

\begin{equation}
H(P,E) = w_1 P + w_2 E,
\end{equation}

where \( H(P,E) \) signifies health intervention effectiveness, \( P \) denotes physiological parameters (e.g., heart rate variability), and \( E \) represents environmental influences (e.g., ambient noise). This capability enables healthcare providers to deliver timely and contextually relevant health prompts that enhance patient outcomes.

In educational contexts, the adaptive gamification subsystem can be employed to create engaging learning environments that respond to student performance metrics. The customization of game mechanics for individual profiles can be expressed as:

\begin{equation}
G_i = \mathcal{F}(P_i, S_i),
\end{equation}

where \( G_i \) denotes the game mechanics for student \( i \), \( P_i \) is performance data derived from assessments, and \( S_i \) is sentiment analysis output generated through Natural Language Processing (NLP). This approach not only fosters motivation but also enhances learning retention by aligning educational content with individual engagement levels.

The business potential of this solution extends beyond immediate productivity gains; it encompasses long-term benefits such as improved employee retention rates, enhanced organizational culture, and reduced healthcare costs. By systematically integrating these advanced methodologies into existing frameworks, organizations can leverage AI to create adaptive environments that promote well-being while driving performance. The strategic application of these technologies positions businesses at the forefront of innovation in workforce management and health optimization.

\section{Results:Metrics for Evaluation}
The evaluation of the proposed AI-driven framework necessitates the establishment of robust quantitative metrics to measure improvements in productivity and employee well-being. Key performance indicators (KPIs) will include employee satisfaction surveys, productivity metrics, and health indicators. Employee satisfaction will be assessed through validated survey instruments that capture dimensions such as job satisfaction, engagement levels, and perceived stress, allowing for a comprehensive understanding of psychological well-being. The survey can be modeled as:

\begin{equation}
S = \frac{1}{N} \sum_{i=1}^{N} s_i,
\end{equation}

where \( S \) represents the overall satisfaction score, \( N \) is the number of respondents, and \( s_i \) denotes individual satisfaction ratings.

Productivity metrics will encompass quantitative measures such as task completion rates, output quality, and efficiency ratios. These metrics can be analyzed using performance data collected from AI systems that monitor employee activities in real-time. The productivity improvement can be expressed as:

\begin{equation}
P = \frac{O_f - O_i}{O_i} \times 100,
\end{equation}

where \( P \) represents the percentage change in productivity, \( O_f \) is the final output after AI intervention, and \( O_i \) is the initial output prior to the implementation.

Health indicators will include physiological parameters obtained from biometric feedback systems, such as heart rate variability (HRV), which serves as a proxy for stress levels. The analysis of health indicators can utilize statistical methods such as regression analysis to identify correlations between AI interventions and health outcomes. The regression model can be formulated as:

\begin{equation}
H = \alpha + \beta_1 P + \beta_2 S + \epsilon,
\end{equation}

where \( H \) denotes health outcomes, \( P \) represents productivity metrics, \( S \) indicates satisfaction scores, \( \alpha \) is the intercept, \( \beta_1 \) and \( \beta_2 \) are coefficients reflecting the relationship strengths, and \( \epsilon \) is the error term.

Statistical analysis will employ techniques such as Analysis of Variance (ANOVA) to compare means across different groups exposed to varying levels of AI interventions. Additionally, multivariate analysis will be utilized to assess the impact of multiple independent variables on dependent outcomes simultaneously. This comprehensive approach to data collection and analysis ensures a rigorous evaluation of the effectiveness of AI-driven solutions in enhancing workplace productivity and promoting employee well-being.

\section{Scope of further Research}
The integration of Artificial Intelligence (AI) within workplace environments is poised to yield significant long-term implications over the next decade. As organizations increasingly adopt AI-driven frameworks, we anticipate a paradigm shift in operational efficiency and employee engagement. The continuous evolution of AI technologies, coupled with advancements in machine learning algorithms, will facilitate more sophisticated models that can dynamically adapt to varying workplace conditions. These models may incorporate enhanced neuroeconomic principles, allowing for real-time adjustments based on employee cognitive and emotional states. The potential for AI to analyze vast datasets will enable organizations to derive actionable insights that inform strategic decision-making processes, thereby fostering a culture of data-driven management.

Scalability remains a critical consideration for the proposed AI system. The framework is designed to be adaptable across diverse industries and organizational sizes, from small startups to large multinational corporations. By employing modular architecture, the system can be tailored to meet specific industry needs, whether in healthcare, finance, or manufacturing. This adaptability can be expressed mathematically as:

\begin{equation}
S = f(I, O),
\end{equation}

where \( S \) represents scalability, \( I \) denotes industry-specific requirements, and \( O \) indicates organizational size and structure. Furthermore, the use of cloud-based infrastructures will facilitate the deployment of AI solutions at scale, enabling real-time data processing and analytics across geographically dispersed teams.

In summary, the long-term implications of AI integration in workplace settings suggest a transformative impact on productivity and employee well-being. The scalability of the proposed system across various sectors underscores its potential to enhance operational effectiveness while addressing unique organizational challenges. Future research should focus on empirical validation of these frameworks through longitudinal studies that assess their effectiveness in real-world applications.

\end{document}